\documentclass[sigconf]{acmart}

\AtBeginDocument{%
  }

\setcopyright{none}
\renewcommand\footnotetextcopyrightpermission[1]{}
\usepackage{listings}
\usepackage{xcolor}
\usepackage{booktabs}
\usepackage{tikz}
\usetikzlibrary{arrows.meta,positioning,shapes.geometric,fit,calc,backgrounds}
\usepackage{xspace}
\usepackage{enumitem}
\usepackage{algorithm}
\usepackage{algpseudocode}

\newcommand{\eg}{\emph{e.g.,}\xspace}

\newcommand{\ours}{Ours}

\begin{document}

\title{LLM-as-Code: Agentic Programming for Agent Harness}

\author{Junjia Qi}
\authornote{Equal contributions.}
\affiliation{%
  \institution{City University of Hong Kong}
  \country{China}}
\email{junjiaqi2-c@my.cityu.edu.hk}

\author{Zichuan Fu}
\authornotemark[1]
\affiliation{%
  \institution{City University of Hong Kong}
  \country{China}}
\email{zc.fu@my.cityu.edu.hk}

\author{Jingtong Gao}
\affiliation{%
  \institution{City University of Hong Kong}
  \country{China}}
\email{jt.g@my.cityu.edu.hk}

\author{Wenlin Zhang}
\affiliation{%
  \institution{City University of Hong Kong}
  \country{China}}
\email{wl.z@my.cityu.edu.hk}

\author{Hanyu Yan}
\affiliation{%
  \institution{City University of Hong Kong}
  \country{China}}
\email{hanyuyan4-c@my.cityu.edu.hk}

\author{Xian Wu}
\affiliation{%
  \institution{Tencent Jarvis Lab}
  \country{China}}
\email{kevinxwu@tencent.com}

\author{Xiangyu Zhao}
\affiliation{%
  \institution{City University of Hong Kong}
  \country{China}}
\email{xianzhao@cityu.edu.hk}

\begin{abstract}
Every major LLM agent framework gives the LLM the role of orchestrator; the model decides what to do next, when to call tools, and when to stop.
We argue that token explosion, control-flow hallucination, and unreliable completion are not implementation bugs but architectural consequences of assigning the deterministic work of looping, branching, and sequencing to a probabilistic system.
A better prompt or a stronger model cannot guarantee the reliability of the LLM agent.
We therefore propose \textbf{Agentic Programming}, in which the program governs all control flow, and the LLM is itself part of it, an adaptive component we call \textbf{LLM-as-Code} and invoke only where a task calls for reasoning or generation.
Within each call the model keeps full flexibility, but it cannot alter the program's execution path.
With control in the program, the LLM's context is built from the execution history's call tree and forms a directed acyclic graph (DAG). Each call's context length is then determined by its call depth rather than by accumulation over steps.
A case study of computer-use agents shows that the design is practical, not just a theoretical stance, substantially improving the stability of long visual operation sequences. The code is available at \url{https://github.com/FuzzyKuji/OpenProgram}.

\end{abstract}




\maketitle

\renewcommand{\thefootnote}{}%
\footnotetext{Accepted at the KDD 2026 Workshop on Agentic Software Engineering (AgenticSE).}%
\renewcommand{\thefootnote}{\arabic{footnote}}%

\section{Introduction}
\label{sec:introduction}

Most mainstream LLM agent frameworks rest on one design decision they seldom defend; the LLM is put in charge of execution.
Given a goal and a set of tools, the model runs the loop, choosing which tool to call~\cite{toolformer2023}, supplying the arguments, reading the result, and deciding when the task is done.
This reason-and-act loop, introduced by ReAct~\cite{react2023}, has become the standard way to build an agent~\cite{agentsurvey2023,agentsurvey2024} and underlies widely used agent frameworks such as AutoGen~\cite{autogen2023}, OpenHands~\cite{openhands2024}, and MetaGPT~\cite{metagpt2024}.
Across these systems, the LLM is the orchestrator; the surrounding program merely relays its calls, while the model itself decides the control flow.
Its appeal is immediate, since a single prompt replaces pages of orchestration code, and on demonstrations and short tasks the agent often appears to perform well without it.

The difficulty surfaces later, when an agent takes on a long-horizon software engineering task~\cite{swebench2024}, such as a multi-file refactor.
The same three symptoms recur.
The context grows with every step until the window must be truncated, and a root-cause hypothesis from earlier is lost; this is token explosion~\cite{lostinthemiddle2024,sametaskmoretokens2024}.
The agent also hallucinates control flow, reporting the issue resolved before the verifier has run~\cite{llmrules2024,followbench2024}.
Finally, completion is unreliable; swayed by the latest failing test, the agent abandons a correct earlier diagnosis, with nothing to guarantee that the steps it should have followed ever ran~\cite{distraction2023,selfcorrect2024}.
These are the long-horizon coherence failures that recent work on agents has begun to document~\cite{seagentreasoning2026,understandingtraj2025}.

Existing approaches treat each symptom as a separate problem to be fixed.
Better prompts raise the chance that the model follows the intended procedure, but the gains are model-specific and never reach a guarantee~\cite{llmrules2024,followbench2024}.
A longer context window holds more history, yet it raises cost and latency while recall still decays before the window fills~\cite{lostinthemiddle2024,sametaskmoretokens2024}.
Smarter retry logic recovers from some failures, but each retry is itself a sampled decision, so it lowers the per-step error rate without bounding the error over a long run~\cite{reflexion2023}.
None of these makes the agent reliable by construction; each narrows the failure rate without eliminating it, so on a long enough task the failure still compounds.
The three symptoms all arise from putting a probabilistic model in charge of control flow that must be deterministic, which programming languages already provide with perfect reliability.

We therefore argue that the LLM should not be the orchestrator; the program should control the flow, and the LLM should reason only when it is called.
We call this alternative \textbf{Agentic Programming}.
The LLM is part of that program, an adaptive component we call \textbf{LLM-as-Code}, invoked only where a task calls for reasoning or generation. Within each call it reasons freely, but it cannot alter the program's execution path.
The paradigm has four parts. These are a code-driven workflow that runs the required steps in order rather than sampling them; a DAG-structured context that stays bounded by call depth instead of growing with every step; multi-agent collaboration that follows for free, since agents are functions over that graph; and self-programmed evolution that commits the agent's improvements as durable code.

Section~\ref{sec:arguments} argues why the orchestrator role is structurally bound to fail, and Section~\ref{sec:solution} presents the LLM-as-Code design and draws out its consequences for software engineering practice.

\section{Why LLM Orchestrator Fails}
\label{sec:arguments}

We argue the failure has three layers.
Section~\ref{subsec:category-error} states the abstract mismatch, since assigning deterministic workflow steps to a probabilistic system is a category error, not an engineering shortfall.
Section~\ref{subsec:control} and Section~\ref{subsec:context} make the error concrete by tracing the two dimensions in which production agents actually break, where compliance cannot be guaranteed and context overflows.

\subsection{Deterministic vs. Probabilistic Workflow}
\label{subsec:category-error}

Typically, the execution of a workflow can be divided into two kinds of step that differ in what correctness requires:
\begin{itemize}[leftmargin=*]
    \item \textbf{Deterministic}: looping, branching, sequencing, variable binding, error handling, all steps whose correctness depends on executing exactly as specified.
    \item \textbf{Probabilistic}: understanding natural language, summarizing text, generating code, making judgment calls, all steps where the ``correct'' output is underdetermined and requires reasoning under uncertainty.
\end{itemize}

Programming languages solve the first kind perfectly; a \texttt{for} loop iterates exactly $n$ times, an \texttt{if} branch is taken if and only if its condition holds, and a function call transfers control and returns.
These guarantees are unconditional; they do not depend on input length, task complexity, or how well the instructions are phrased.
LLMs, by contrast, are designed for the second kind; their strength is producing useful outputs for inputs they have not seen before.
That same strength entails a structural weakness, because every output is sampled from a distribution, including the outputs that encode control decisions; asking an LLM to decide ``what to do next'' is sampling a control-flow edge from a probability distribution.

\begin{figure*}[!t]
\centering
\includegraphics[width=0.8\linewidth]{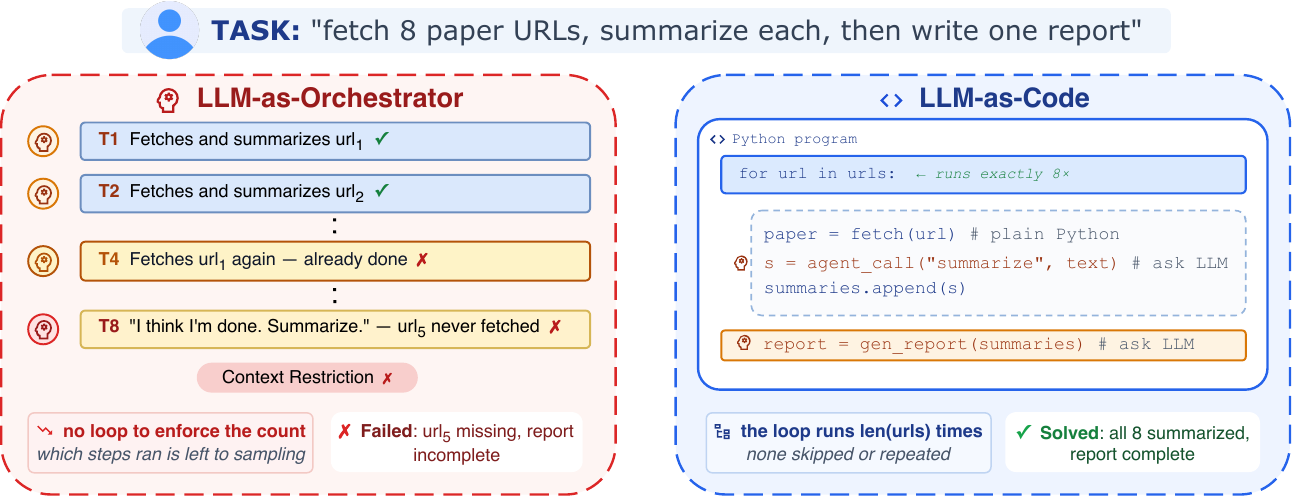}
\caption{A comparison of the two agent paradigms on a simple multi-step task. In the left panel (LLM-as-Orchestrator) the LLM drives the loop itself. In the right panel (LLM-as-Code) code owns the loop and the LLM is called only to summarize.}
\Description{A diagram comparing two agent architectures side by side, on the task of fetching eight URLs, summarizing each, and writing one report. The left panel (LLM-as-Orchestrator) shows the model running the loop itself: it fetches and summarizes url1 and url2, but with no loop variable to track progress it re-fetches url1, never fetches url5, and declares completion early, leaving the report incomplete. The right panel (LLM-as-Code) shows a \texttt{for url in urls} loop that fetches each URL exactly once and calls the LLM only to summarize each page and to write the final report, so all eight URLs are processed.}
\label{fig:comparison}
\end{figure*}

Figure~\ref{fig:comparison} makes this concrete with a small task that fetches eight URLs, summarizes each, and produces a report.
Under LLM-as-Orchestrator (left), the model itself decides, at each turn, which URL to fetch next, whether to summarize now or fetch more, and when to write the report.
The task demands no sophisticated planning; a three-line \texttt{for} loop discharges it (Figure~\ref{fig:comparison}, right).
When the model drives the loop itself, however, it re-fetches URL\textsubscript{1}, omits URL\textsubscript{5}, or terminates prematurely (Figure~\ref{fig:comparison}, left).

The failure is structural rather than a limitation of model capability: correct execution requires deterministic iteration, a guarantee that sampling cannot furnish.
Because every control decision is drawn from a distribution whose per-step accuracy is below one, the probability of a fully correct run diminishes as the horizon lengthens, and a stronger model reduces the per-step error without eliminating its compounding; the only construction that guarantees a loop executes $n$ times is to write the loop.
The underlying error is that LLM-as-Orchestrator delegates the deterministic operations of iteration, branching, sequencing, and termination to a probabilistic system: control-flow hallucination and unreliable completion follow directly, and token explosion and context overload follow once execution history must be retained as a single flat list with no locus for compaction.
These pathologies persist across models, frameworks, and prompting strategies because they originate in the architecture rather than the implementation.

\subsection{Unguaranteed Compliance}
\label{subsec:control}

The same category error surfaces a second way when the workflow has structure that must be enforced, such as a check that must run before a write.
Every control decision is sampled, so the orchestrator has no native means to obey a constraint as a constraint.
At best it is told about the constraint in the prompt and asked to follow it.
Whether it does so is a probabilistic outcome.

For short, simple workflows this is often acceptable, since a capable model will follow a three-step recipe most of the time.
But realistic agent skills are not short or simple.
Consider a single skill bundling twenty-plus rules governing concerns such as file naming and write ordering.
The model is asked to generate behavior that simultaneously satisfies all of them.
Empirically, it does not~\cite{llmrules2024,followbench2024}; some rules are attended to while others are forgotten by step five, and conflicts are resolved by sampling rather than by precedence.
Adding more rules does not monotonically improve compliance; past a point, every new rule competes for attention with the others, and the agent's behavior becomes harder to reason about, not easier.

The only construction that enforces a workflow's rules is one in which the rules are not asked of the model at all.
A \texttt{for} loop runs $n$ times because the language runtime guarantees it; an \texttt{if} branch is taken iff its condition holds; a function returns before its caller resumes.
What the LLM is asked to do in such a workflow shrinks to the few decision points where reasoning is genuinely needed, such as generating a summary or judging whether a check passed.
The rest of the workflow is not begged from the model; it is executed.
From the standpoint of workflow control, then, LLM-as-Orchestrator is structurally weak; it can be coached toward compliance, but it cannot guarantee compliance, and the harder the workflow the more the gap matters.

\subsection{Context Overflow}
\label{subsec:context}

If the LLM owns scheduling, the entire history must be re-shown each turn, including every tool call, output, and reasoning trace, so the model can re-derive ``where am I in the task.''
Context therefore grows as $O(\text{steps} \times \text{avg\_output})$, unbounded by construction on a long-horizon task.
The cost is twofold; a hard limit is enforced by the window, and a soft degradation appears well before that limit.

\paragraph{Hard limit: the window does not fit.}
Every model has a maximum context length.
When the conversation crosses it, the system must either truncate or summarize, and both destroy information irreversibly, where the lost information may include an earlier root-cause hypothesis or a constraint mentioned twenty steps ago and now invisible.
The orchestrator wakes up the next turn with a redacted view of its own history.
A bigger window or smarter compaction postpones this point but does not change its inevitability; the window can be enlarged, but its structure, a flat conversation log re-fed every turn, is wrong for a branching execution trace.

\paragraph{Soft cost: reasoning degrades long before the window fills.}
Even inside the window, longer contexts hurt the LLM in two ways.
Reasoning quality drops; long-context recall is empirically weaker, attention to early information thins, and accuracy on multi-step tasks degrades as the prompt grows~\cite{lostinthemiddle2024,sametaskmoretokens2024}.
Throughput drops too; each additional turn is more expensive in tokens and latency, so a long task progressively slows and costs more per step.
By step fifty, the orchestrator is reasoning over its own accumulated transcript, slowly, expensively, and worse than it would have on the same problem at step ten.

\section{Agentic Programming}
\label{sec:solution}

To make LLM agents reliable despite the failures identified in Section~\ref{sec:arguments}, we propose \textbf{Agentic Programming}, a programming paradigm in which the program drives execution and the LLM is part of the program itself.
The control flow is written in ordinary code, and the LLM serves as an adaptive component invoked where a task calls for reasoning or generation.
Instantiated for an agent, this paradigm yields the relationship we call \textbf{LLM-as-Code}.
We develop it in four parts. These are a code-driven agent workflow (Section~\ref{subsec:spine}), a DAG-structured context (Section~\ref{subsec:graph-context}), multi-agent collaboration (Section~\ref{subsec:multiagent}), and self-programmed evolution (Section~\ref{subsec:benefits}).

\subsection{Code-driven agent workflow}
\label{subsec:spine}
In Agentic Programming, the agent's workflow is formed with code, so the program controls execution and calls the LLM only at the steps that need reasoning.
This sits between two extremes, where a hardcoded program runs exactly what was written and nothing else, while an LLM-orchestrated agent can do anything but guarantees nothing, since every step rides on a sampled decision.
Agentic Programming gives each its own layer.
The workflow lives in the program, where the runtime enforces each rule rather than asking the model to follow it.
At every node of that workflow the LLM is unrestricted, free to use any tool, prompt, or skill the task calls for.
What we promise about a workflow is delivered by the program, not begged from the model.
This is the direct answer to Section~\ref{subsec:control}; compliance is no longer sampled, since the rules are never asked of the model at all.

The workflow is not a fixed pipeline.
When an LLM judgment calls for more work, the function it invokes is itself ordinary code that wraps its own LLM calls, so the call graph is unfolded at runtime rather than fixed in advance (Appendix~\ref{app:agentic-call} describes the call-site decorator and runtime in more detail).
A hardcoded pipeline fixes both the graph's shape and the logic at each node, and a workflow engine such as Airflow or Temporal likewise fixes the composition before execution; LLM-as-Orchestrator instead surrenders both to sampling.
Agentic Programming fixes only the discipline of each layer and lets the LLM's judgments shape the graph. An orchestrating function can ask ``which stage next?'' and branch on the answer through a switch in the program, while an open-ended task such as ``explore this code for security issues'' runs as a bounded loop whose \texttt{max\_steps} and stopping checks are the program's and whose per-step decisions are the LLM's.
Flexibility therefore lives in the recursive depth and reliability in each layer's code; neither is traded for the other.

\subsection{DAG-structured context}
\label{subsec:graph-context}

\begin{figure*}[!t]
\centering
\includegraphics[width=0.85\linewidth]{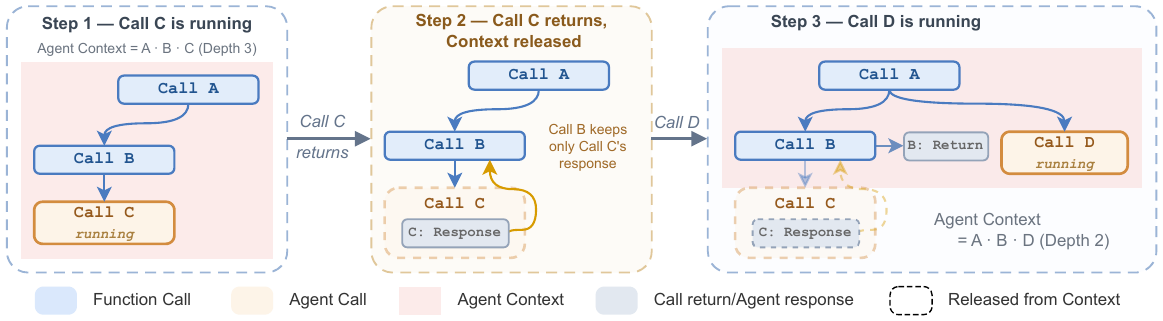}
\caption{Context tracks the call graph rather than a flat conversation log. A running call keeps its full ancestor chain, while a call that has returned collapses to a one-line summary, so as execution moves from $C$ to its sibling $D$ the live context stays bounded by the current call depth, not the total number of steps. See Section~\ref{subsec:graph-context}.}
\Description{A three-step trace of context as call-graph scope on the call tree where A calls B, B calls C, and D is a second child of A. Step 1: call C is running and the live agent context is the full chain A, B, C at depth 3. Step 2: call C returns and its frame is released, so its parent B keeps only C's one-line response in place. Step 3: the sibling call D is running with context A and B in full plus C as a single summary line, at depth 2, while C is folded out of context. A legend distinguishes function calls, agent calls, the agent-context region, returned values and agent responses, and frames released from context.}
\label{fig:context}
\end{figure*}

Once the agent's workflow is driven by code, its execution history stops being a flat conversation log and becomes the call graph itself, a directed acyclic graph (DAG) of function and reasoning nodes; in the simple case this graph is a tree, becoming a DAG once parallel branches merge back.
Context then behaves like ordinary function-call memory instead of an ever-growing transcript, governed by a single scoping rule.
\emph{A call sees the full context of its entire ancestor chain, while each frame keeps only a summary of the children that have already returned.}

Figure~\ref{fig:context} traces the rule over three steps on the call tree $A \to \{\,B \to C,\ D\,\}$, where $A$ and $B$ are function calls and $C$ and $D$ are reasoning calls.
While $C$ runs, the live context is the full chain $A, B, C$ at depth three.
When $C$ returns, its frame is released and its parent $B$ keeps only $C$'s one-line summary in its place.
When the sibling $D$ runs next, it inherits $B$'s returned context, so its input is $A$ and $B$ in full plus that single summary of $C$, now at depth two, never $C$'s prompts or intermediate work.
Where parallel branches merge, their results are likewise compressed before passing upward.

This answers Section~\ref{subsec:context} on both counts.
The hard limit is relieved because no call ever carries the whole task's history, only its ancestor chain, with already-returned subtrees collapsed to summaries.
Input length is therefore bounded by the current depth, $O(\text{depth})$, not by the total number of steps run, $O(\text{steps})$, the way a flat log that grows monotonically and never shrinks would be.
The soft degradation is relieved because each call reasons in a short, scoped prompt, free of the deliberation beside and below it.
Nothing is discarded; the full DAG is retained at the harness level for replay and debugging, even though no node ever sees all of it.
The structure, state, and execution-grounding that recent work asks agents to bolt on~\cite{seagentreasoning2026} arrive here for free, supplied by the call graph and the real execution results its return values carry rather than added as a separate mechanism (Appendix~\ref{app:patches}).

\subsection{Multi-agent collaboration}
\label{subsec:multiagent}
An agentic function is an ordinary function, so running several agents is invoking several functions, which the program executes in parallel as it would any other code.
Each agent reasons within its own call, so independent model invocations proceed concurrently rather than through a single controller, and the DAG-structured context of Section~\ref{subsec:graph-context} carries over without change: concurrent agents are sibling branches of one call graph, each scoped to its own path, their outputs merged where the branches rejoin.
Coordination is therefore the graph itself rather than a supervisory model, and the sampled control whose unreliability Section~\ref{subsec:control} establishes is never reintroduced at the level of the collective.
Two consequences matter in practice: scale is bounded by resources rather than by a shared context window, since each agent returns only a summary and the parent accumulates results rather than transcripts; and the interface between agents is a typed return value rather than a free-form conversation, so a failed agent is a failed call that the program may retry, substitute, or abandon under its own rules.

A code-review agent makes the structure concrete.
It dispatches one sub-agent per modified file, each reviewing within its own scoped context, and the parent consolidates their findings once all have returned.
The cost of the fan-out is legible in the call graph: a single orchestrator reviewing every file in one context would carry a prompt that grows with the combined length of all reviews, whereas here the parent retains one summary per file and the depth of any individual review never exceeds that file's own subtree.
The join is program logic rather than a delegated judgment: the parent deduplicates findings that recur across files, orders them by severity, and decides what to surface, so the consolidation is reproducible and inspectable in a way a model-mediated merge is not.
Failure is localized in the same way, since a sub-agent that stalls or returns an ill-formed result is one failed call that the program may retry, route to a stronger model, or record as an unreviewed file, without perturbing the others.

\subsection{Self-programmed evolution}
\label{subsec:benefits}
An agentic function is just a function, so the paradigm composes with itself. A function that generates and refines other agentic functions runs under the same discipline, its loop and acceptance criteria the program's and its judgments LLM calls.
This self-evolution belongs to the deterministic layer of Section~\ref{subsec:category-error}. Proposing a new function is a probabilistic LLM call, but the result is committed as code (accepted only once it passes the caller's tests) and thereafter runs like any other guaranteed step~\cite{metagpt2024}.
A self-improving orchestrator instead evolves as text re-read and re-decided each run, so whether a lesson is followed is itself sampled (Section~\ref{subsec:control}); here the agent revises what it does, not just what it knows, and the revision is durable.
The same footing restores ordinary engineering practice. A unit test can pin an LLM reply and assert deterministically on the enclosing function's output, existing software can adopt LLM reasoning one call site at a time rather than be rewritten as an agent, and a failure localizes to a named call's return value rather than a single turn within a fifty-step transcript.

\begin{table}[t]
\centering
\caption{GUI automation on OSWorld~\cite{osworld2024} (overall success \%). Baseline numbers are from its public leaderboard (accessed 2026-06-02) and run up to 100 steps; \ours{} wins in 15 steps.}
\label{tab:osworld-body}
\begin{tabular}{l c c}
\toprule
\textbf{Method} & \textbf{Max Steps} & \textbf{Overall} \\
\midrule
Holo3-35B-A3B   & 100 & 80.4 \\
OpenAPA w/ Gemini-3.1-pro     & 100 & 78.3 \\
Claude Sonnet 4.6   & 100 & 72.1 \\
\midrule
\textbf{LLM-as-Code} w/ Claude Sonnet 4.6                & \textbf{15} & \textbf{86.8} \\
\bottomrule
\end{tabular}
\end{table}

Agentic Programming does not subsume every agent.
Fully exploratory tasks with no known structure, such as open-ended brainstorming or research without a stage model, may benefit from LLM-driven orchestration, which we do not contest.
The claim is narrower. When a workflow has a known structure, that structure should be written as a program, not delegated to the model.
On exactly such structured tasks, two production agents built on this design bear the claim out.
Table~\ref{tab:osworld-body} reports our GUI agent on OSWorld~\cite{osworld2024}, where every baseline is allotted 100 steps and the strongest still tops out at 80.4, the step-burning Section~\ref{subsec:category-error} predicts of a sampled controller, while our agent runs the loop in code and reaches 86.8 in 15 steps.
Appendix~\ref{sec:evidence} reports both in full.

\section{Conclusion}

The dominant execution model in LLM agent frameworks commits a category error, since it assigns the deterministic work of looping, branching, and sequencing to a probabilistic system, so the failures are architectural and will not be resolved by scaling alone.
We move control out of the model, so the program owns execution while the LLM handles reasoning, generation, and judgment, invoked recursively so flexibility lives in the call graph and reliability in each layer's code.
As LLM agents become standard in software engineering, the question is not what the LLM can do but who should control execution. This responsibility belongs with the program, which has long discharged it reliably.

\bibliographystyle{ACM-Reference-Format}
\bibliography{9References}

\appendix
\section{Why Standard Patches Miss}
\label{app:patches}

The category error is not for lack of effort, since the community has produced a rich line of patches.
Each, we argue, treats one symptom while leaving the underlying mismatch in place.

\paragraph{Bigger context windows and smarter compaction.}
Extending the window or summarizing old turns makes the flat conversation log \emph{larger} or \emph{lossier}, but it is still a flat log.
The representation remains wrong for a branching execution; history is re-sent every round, and eviction still discards information irreversibly.
Enlarging the log does not change the fact that its representation is wrong.

\paragraph{Self-reflection and retry (ReAct~\cite{react2023}, Reflexion~\cite{reflexion2023}).}
Reflection lowers the per-step error rate, but control is still \emph{sampled}, since the model decides whether to retry, what to retry, and when to stop.
Lowering per-step error does not yield a multi-step guarantee, since the failure probability still compounds with task length (Section~\ref{subsec:category-error}).

\paragraph{Constrained and structured decoding.}
Forcing the next action into a valid schema guarantees the \emph{form} of a control decision, not its \emph{correctness}.
One can compel well-formed JSON for ``next step'' yet still emit a sequence that skips a step or terminates early; the schema constrains syntax, not the trajectory.

\paragraph{Code as the action representation (\eg CodeAct~\cite{codeact2024}).}
A concurrent line of work argues that letting the LLM emit and execute code as its action, rather than as structured tool calls, produces stronger agents.
We agree that code is the right action representation but go one layer deeper, since code should not only be \emph{the LLM's action}, it should \emph{own the loop the LLM acts within}.
That work makes each step a code action; we make the program own the sequence of steps.

\paragraph{Plan-and-execute.}
Splitting planning from execution is a step toward structure, but the plan is still LLM-generated and therefore still sampled.
A wrong or incomplete plan is executed faithfully; control has been delegated to a probabilistic artifact one level up, not removed from the probabilistic system.

\paragraph{Code-defined state graphs (\eg LangGraph~\cite{langgraph2024}).}
LangGraph comes closest in implementation; the graph is defined in Python and the LLM is invoked at its nodes, so control already lives in code.
Two differences remain.
Its graph is \emph{pre-declared}, with nodes and edges fixed before the run, whereas ours is unfolded by recursive call at runtime.
And its context is threaded through a shared state object that every node reads and writes, whereas ours is released by stack unwinding, so a returned branch leaves nothing behind but its return value.
LangGraph composes a state machine the author draws; Agentic Programming composes a call graph the program grows.

\paragraph{Prompt-compiled pipelines (\eg DSPy~\cite{dspy2024}).}
DSPy is the closest in spirit, where Python composes the modules, and an optimizer tunes prompts and demonstrations against examples.
The complement of focus is instructive. DSPy optimizes \emph{how} each call performs within a graph the author fixes at compile time, while Agentic Programming concerns \emph{who} fixes that graph in the first place, and lets the LLM's runtime judgments shape it.
DSPy improves what each call produces; we ask who owns the flow.

\paragraph{Adding structured state to the orchestrator.}
The closest position to ours is also the most instructive contrast.
A recent position paper~\cite{seagentreasoning2026} diagnoses today's SE agents as fundamentally \emph{reactive}, deciding from conversation history and the latest response with no explicit structure or persistent state, and calls for reasoning that is \emph{structured}, \emph{state-aware}, and \emph{execution-grounded}.
We share the diagnosis entirely, but its remedy attaches an evolving state to an agent that is \emph{still the orchestrator}.
As long as the LLM owns control, that state is read and updated by a probabilistic loop, so the very forces that produce drift, such as overfitting to the latest output, now corrupt the state instead of the transcript.
An orchestrator with an explicit state still decides, stochastically, what to record in it.
The structure, state, and grounding it asks for follow more cleanly from taking orchestration away from the model than from giving it a memory (Section~\ref{subsec:graph-context}).

The pattern is consistent, since every patch improves the \emph{substance} of reasoning or shrinks a symptom, while leaving \emph{control of the flow} inside the probabilistic system.
The category error is addressed only by moving control out.

\section{An Example Skill}
\label{app:skill-example}

To make Section~\ref{subsec:control}'s ``twenty-plus rules'' concrete, the listing below sketches a (lightly anonymised) skill from one of our production agents.
The skill specifies how the agent must add a new feature to a Python codebase.

\begin{lstlisting}[language=,basicstyle=\ttfamily\footnotesize,xleftmargin=4pt,morekeywords={}]
SKILL_FEATURE_ADDITION = [
    # naming
    "snake_case for functions, PascalCase for classes",
    "test files mirror sources: src/foo.py -> tests/test_foo.py",
    # ordering (must hold throughout the task)
    "run the linter before any commit",
    "run the test suite locally before pushing",
    "rebase on main before opening a PR",
    "open a draft PR before requesting review",
    # safety
    "no print() in production; use the logger",
    "every public function requires type hints",
    "every public function requires a docstring",
    "every new public function gets a unit test",
    # change discipline
    "modify at most three files per commit",
    "commit messages start with feat:, fix:, or chore:",
    # exit
    "verify CI green before merging",
    "delete the feature branch after merge",
]
\end{lstlisting}

\noindent The orchestrator must satisfy all of these simultaneously, on every step, for the duration of the task.
Empirically~\cite{llmrules2024,followbench2024}, compliance degrades as the rule set grows, and the rules forgotten first tend to be the procedural ones, such as \emph{run the linter}, rather than the naming ones. Procedural rules require remembering ``what step am I at,'' which is exactly the decision LLM-as-Orchestrator samples rather than enforces.

\section{Agentic Function Calling}
\label{app:agentic-call}

Each LLM call site in our reference implementation is a Python function marked with a decorator.
The decorator hides the prompt template, the model, and the input/output schema behind the function's signature, so the surrounding program invokes the LLM exactly the way it invokes any other function.

\begin{lstlisting}[language=Python]
@agentic
def summarize(text: str) -> str:
    """Summarize the text in 2-3 sentences."""

@agentic
def choose_next_stage(task: str,
                      options: list[str]) -> str:
    """Pick the next stage to tackle from the options."""
\end{lstlisting}

The body of an \texttt{@agentic} function is just a docstring, with no Python code to run inside it.
At call time, the runtime takes the function's signature and docstring, fills a prompt template with the typed arguments, calls the underlying model, and parses the response into the declared return type before handing it back to the caller.
The call site then looks like an ordinary function call.

\begin{lstlisting}[language=Python]
@agentic
def fix_issue(repo: Repo, issue: Issue) -> Patch:
    """Produce a patch that closes the issue."""

def run(repo, issues):
    patches = []
    for issue in issues:
        patch = fix_issue(repo, issue)   # calls LLM
        if patch.verifies(repo):         # plain Python
            patches.append(patch)
    return patches
\end{lstlisting}

\emph{Agentic function calling} is then the natural composition pattern.
An \texttt{@agentic} function with a complex declared task, for instance ``produce a patch that closes this issue,'' can be implemented at runtime by a body that itself calls other \texttt{@agentic} functions, such as a verification step. The runtime serves each of these through the same decorator machinery.
Composition lives in the surrounding Python, not in the prompts; the call graph is unfolded turn by turn rather than declared up front.

\section{A Self-Evolution Walkthrough}
\label{app:self-evolution}

The walkthrough below shows how an \texttt{@agentic} function can be regenerated at runtime when its tests start failing, illustrating Section~\ref{subsec:benefits}'s claim that self-evolution under our paradigm hardens into the deterministic layer rather than living as text the model must re-read.

\begin{algorithm}[H]
\caption{Self-evolution of an agentic function}
\label{alg:self-evolution}
\begin{algorithmic}[1]
\Require an agentic function $f$ with a test suite $T$
\Require a meta-function $g$ that proposes new agentic functions
\Loop
    \If{$f$ passes every test in $T$}
        \State \textbf{continue}
    \EndIf
    \State $f' \gets g(f,\ \text{tests that } f \text{ failed})$ \Comment{LLM call: propose a replacement}
    \If{$f'$ passes every test in $T$}
        \State $f \gets f'$ \Comment{commit $f'$ as code}
    \EndIf
\EndLoop
\end{algorithmic}
\end{algorithm}

Two properties are worth noting.
First, the only sampling step is line 6 (proposing $f'$); the acceptance check on line 7 is deterministic.
Second, line 8 commits the replacement \emph{as code}, so from that point onward, every invocation of $f$ runs $f'$ unconditionally, the way any other Python function runs.
Under an orchestrator, by contrast, the new behaviour would have to be carried as a remembered note that the model must re-read and choose to apply on each subsequent turn, with the no-guarantee semantics of Section~\ref{subsec:control}.

\section{Empirical Evidence}
\label{sec:evidence}

The case for Agentic Programming does not rest on argument alone.
We have built two agents under this paradigm, a GUI automation agent and an autonomous overnight research pipeline.
Both share the same structure, where loops own iteration and termination, while LLM calls reason only at the leaves. Each agent's top-level loop fits in roughly 150 lines of Python.

\begin{table}[!t]
\centering
\caption{Success rate (\%, higher is better) on the three OSWorld~\cite{osworld2024} domains our GUI automation agent targets, comparing \ours{} against agentic frameworks, general models, and task-specialized models. Baseline numbers are from the OSWorld leaderboard (accessed 2026-06-02). See Section~\ref{sec:evidence}.}
\label{tab:osworld}
\vspace{4pt}
\small
\resizebox{\linewidth}{!}{%
\begin{tabular}{l l c c c c c}
\toprule
\textbf{Type} & \textbf{Method} & \textbf{Max Steps} & \textbf{Chrome} & \textbf{Multi-Apps} & \textbf{OS} & \textbf{Overall} \\
\midrule
Agentic framework & Agent~S3~\cite{agents32025} w/ Opus 4.5 bBoN (N=1) & 100 & 61.7 & 57.4 & 75.0 & 66.0 \\
Agentic framework & UiPath Screen Agent w/ Opus 4.5 & 100 & 69.5 & 53.0 & 70.8 & 67.1 \\
Agentic framework & Agent~S3 w/ GPT-5 bBoN (N=10) & 100 & 69.5 & 58.7 & 79.2 & 69.9 \\
General model & claude-sonnet-4-6 & 100 & 78.5 & 60.2 & 91.7 & 72.1 \\
Agentic framework & Agent~S3 w/ Opus 4.5 + GPT-5 bBoN (N=10) & 100 & 63.0 & 63.9 & 79.2 & 72.6 \\
General model & Kimi K2.6 & 100 & 76.7 & 55.0 & 79.2 & 73.1 \\
Agentic framework & HIPPO Agent w/ Opus 4.5 & 100 & 60.4 & 64.3 & 87.5 & 74.5 \\
Agentic framework & VLAA-GUI w/ Opus 4.5 & 100 & 66.6 & 61.1 & 91.7 & 76.3 \\
Agentic framework & OpenAPA w/ gemini-3.1-pro & 100 & 71.7 & 65.7 & 83.3 & 78.3 \\
Specialized model & Holo3-35B-A3B & 100 & 78.3 & 62.9 & 95.8 & 80.4 \\
\midrule
\textbf{Agentic framework} & \textbf{\ours} & \textbf{15} & \textbf{93.5} & \textbf{80.0} & \textbf{100.0} & \textbf{86.8} \\
\bottomrule
\end{tabular}%
}
\end{table}

Table~\ref{tab:osworld} reports the GUI agent on the OSWorld benchmark~\cite{osworld2024}.
It reaches the best result on all three domains it targets and on the overall score (86.8 versus 80.4 for the strongest prior system), ahead of agentic frameworks, general models, and task-specialized models alike.
This advantage follows from reliability by construction; the required steps are written in code and always run in the prescribed order, since no sampled decision can skip, repeat, or misorder them, and the LLM is invoked only for the judgments inside those steps, never for the decision to take them.
Beyond the benchmark score, the paradigm reshapes everyday engineering, since failures localize to a named call's return value rather than a turn in a fifty-step transcript, and neither agent was designed for these properties; both inherited them from where control sits.

\end{document}